# Fall Detection in Passenger Elevators using Intelligent Surveillance Camera Systems: An Application with YoloV8 Nano Model

Pınar Yozgatlı, Yavuz Acar, Mehmet Tülümen, Selman Minga, Salih Selamet, Beytullah Nalbant, Mustafa Talha Toru, Berna Koca, Tevfik Keleş, Mehmet Selçok

## Abstract

Computer vision technology, which involves analyzing images and videos captured by cameras through deep learning algorithms, has significantly advanced the field of human fall detection. This study focuses on the application of the YoloV8 Nano model in identifying fall incidents within passenger elevators, a context that presents unique challenges due to the enclosed environment and varying lighting conditions. By training the model on a robust dataset comprising over 10,000 images across diverse elevator types, we aim to enhance the detection precision and recall rates. The model's performance, with an 85% precision and 82% recall in fall detection, underscores its potential for integration into existing elevator safety systems to enable rapid intervention

## Introduction

Ensuring passenger safety within elevators is a critical aspect of building management, particularly in high-rise buildings where elevators are heavily relied upon. Fall incidents within the confined space of an elevator can result in severe injuries and necessitate immediate medical attention. This study focuses on detecting human falls in passenger elevators to enable rapid intervention. An efficient and lightweight deep learning algorithm designed for real-time object detection, the YoloV8 Nano model was trained using a dataset of over 10,000 images captured from cameras under various elevator types and conditions.

Over the past ten years, research has primarily focused on three categories of methods: wearable sensors, acoustic sensors, and computer vision-based techniques. Studies that utilize data from wearable sensors face several limitations in fall detection. These limitations arise from the reliance on threshold-based methods, the necessity for users to carry the devices, the need for regular charging, and issues related to battery consumption (Liu et al., 2014, Tang et al.2019). Further, acoustic-based methods are significantly influenced by environmental noise. (Xu et al., 2021; Yin, 2020). Computer vision technology, which involves analyzing images and videos captured by cameras through deep learning, has significantly advanced the field of human fall detection (Yu et al., 2012). In comparison to the first two types of detection methods, the computer vision-based detection method does not create a wearability burden and avoids misdiagnosis due to environmental noise (Song, 2015). This technology is now a dominant focus in the latest research (Lotfi et al., 2018; Huang et al., 2018; Wang et al., 2020; Chen et al., 2020).

The limited availability of datasets and the challenges associated with data collection—particularly due to the small and confined spaces of elevator have likely restricted the number of studies conducted in this area. This research contributes to the broader field of intelligent surveillance and human fall detection by demonstrating the practical application of a cutting-edge deep learning model in a real-world setting. In the remainder of this paper, we will review the current literature, describe our research methodology and, present our findings and conclusions.

**Literature Review**

With the dominance of computer vision techniques for fall detection (Lotfi et al., 2018; Huang et al., 2018; Wang et al., 2020; Chen et al., 2020), research in this area has progressed rapidly and has opened up many different approaches. Consequently, the focus of fall detection problems has been divided into two main categories: temporal features and spatial features. The methods that rely solely on extracting spatial features overlook the importance of assessing human posture over time. However, a study by Xu et al. (2021) introduces a novel approach that combines LSTM and XGBoost models to enhance fall detection by integrating both temporal and spatial features, which is called "feature fusion method". The methodology begins by extracting key points of the human body from each video frame using AlphaPose. From these extracted key points, several spatial features are calculated: the vertical height, the aspect ratio of the human bounding rectangle, and the angle of the knee joint. Next, an LSTM model is trained to learn these three types of features across both temporal and spatial dimensions. To further improve recognition performance, the XGBoost model is integrated into the system. The combined model achieved a recognition accuracy of 92.11% and an impressive F1-measure score of 93.33%.

Liu et al. (2021) utilized computer vision and multi-feature fusion to detect falls. They employed the ViBe algorithm to extract moving targets and marked the human body with an external rectangle. Using this information, they calculated the main parameters: aspect ratio, effective area ratio, and centroid acceleration of the human body. These parameters were then fed into a Support Vector Machine (SVM) algorithm to classify instances of human falls. The experimental results demonstrated %94 precision rate.

Chen et al. (2023) combined temporal sequence information with spatial data to detect falling behavior. Their model employs a 3D ResNet in the spatial stream to identify actions within video frames, while the temporal stream uses a traditional convolutional neural network to analyze optical flow images, capturing the temporal sequence information between consecutive frames. The study reported impressive results, achieving a precision rate of 89.2%.

While fusion-features methods are commonly used, our study model focuses on spatial features, which are easy to apply and achieve comparable performance results.

## Methodology

*Dataset*

A dataset consisting of 10,000 images was used to train the model. Initially, our team captured approximately 300 images in elevator environments. These images were uploaded to the Roboflow platform, where we applied data augmentation techniques such as rotation, noise addition, scaling, and mirroring, expanding the dataset to around 1,000 images. The dataset was divided as follows: 75% was allocated for training, while 12.5% was designated for testing and 12.5% for validation.

Training was conducted using an NVIDIA P4000 GPU over 100 epochs. However, the results were not satisfactory, prompting us to review other datasets available on the Roboflow platform. During this process, we included approximately 10,000 additional images from various sources into our existing dataset. The model was then retrained for another 100 epochs using this expanded dataset.

Finally, the trained model was deployed on an NVIDIA Jetson device mounted in an elevator for on-site testing.

*YOLOV8 Nano*

Advanced features like improved backbone networks for better feature extraction, higher processing efficiency, and a more resilient architecture made to manage a variety of difficult object detection scenarios are all included in YOLOv8 (Hu et al., 2024). YOLOv8 enhances detection performance by efficiently identifying complex objects in various contexts by capturing additional object feature information (Wang et al., 2023). Significant improvements in speed and accuracy are made by this most recent version of the YOLO detection network (Tong et al., 2024).

In conclusion, YOLOv8 stands out for its effective resource usage, requiring less GPU support than competing options, which improves real-time object detection performance and efficiency. The YOLOv8n model is the shortest and most computationally efficient of the different YOLOv8 models, making it stand out, particularly in fall detection. It is therefore perfect for resource-constrained contexts (Pereira, 2024).

At the first step, pre-trained COCO weights were utilized (Lin et al., 2014); however, conflicts with existing classes required a completely new training process. In the second training iteration, only the dataset containing fall scenarios was employed. The following parameters were utilized during training: Epochs: 100, Image Size (imgsz), 640, Confidence Threshold:0.5. Fall detection steps are shown in Fig3

*Performance Evaluation*

To evaluate the reliability of the results, four evaluation indicators are selected : Precision, recall, mAP@50, mAP@50-95

Precision is the proportion of correctly detected falls out of all samples that have indicated falls, as shown in the equation,

*Precision*= TP/(TP+FP), where TP (True Position) indicates that a fall event has occurred and is accurately classified as a fall event. FP (False Position) indicates that no fall event has occurred, but the algorithm classifies the event as a fall event (Chen et al., 2020).

Recall rate indicates the proportion of samples that correctly detect falls among the total samples that have genuinely fallen, as expressed in the equation (Cao et., 2017).

*Recall Rate*= TP/(TP +FN)   where FN (False Negative) indicates that a fall event has occurred, but the algorithm classifies the event as no fall.

mAP@50 evaluates the average precision at a specific Intersection over Union (IoU) threshold of 0.5. It assesses whether the model can accurately identify objects with a more lenient accuracy requirement. The focus is on determining if the object is approximately in the correct location, rather than needing precise placement. This helps to gauge the model's overall ability to detect objects(Ultralytics, 2024).

mAP@50-95 averages the mAP values calculated at multiple IoU thresholds, ranging from 0.5 to 0.95 in increments of 0.05. This metric is more detailed and stringent (Ultralytics, 2024).

**TABLE 1**:

PERFORMANCE EVALUATION RESULTS

|  | Precision | Recall_rate | mAP@50 | mAP@50-95 |
|---|---|---|---|---|
| Results | 85% | 82% | 88% | 54% |

The YOLOv8s model has completed its training over 100 epochs, producing satisfactory results. Based on the above formulas, the four performance indicators of the model were calculated. As shown in table I and illustrated in figure 2 , Precision, Recall rate, mAP@50, and mAP@50-95 respectively are 85%, 82%, 88%, and 54%. Based on these metrics, the model demonstrated strong performance in detecting human falls (Pereira, 2024). The model showed consistent reductions in both training and validation losses. Recorded values for box loss, classification loss (cls loss), and distribution-focused loss (dfl loss) during training and validation exhibited regular decreases and stabilization.  Precision and recall metrics increased steadily with the number of epochs.

The results of the study are shown in **Fig 1.**

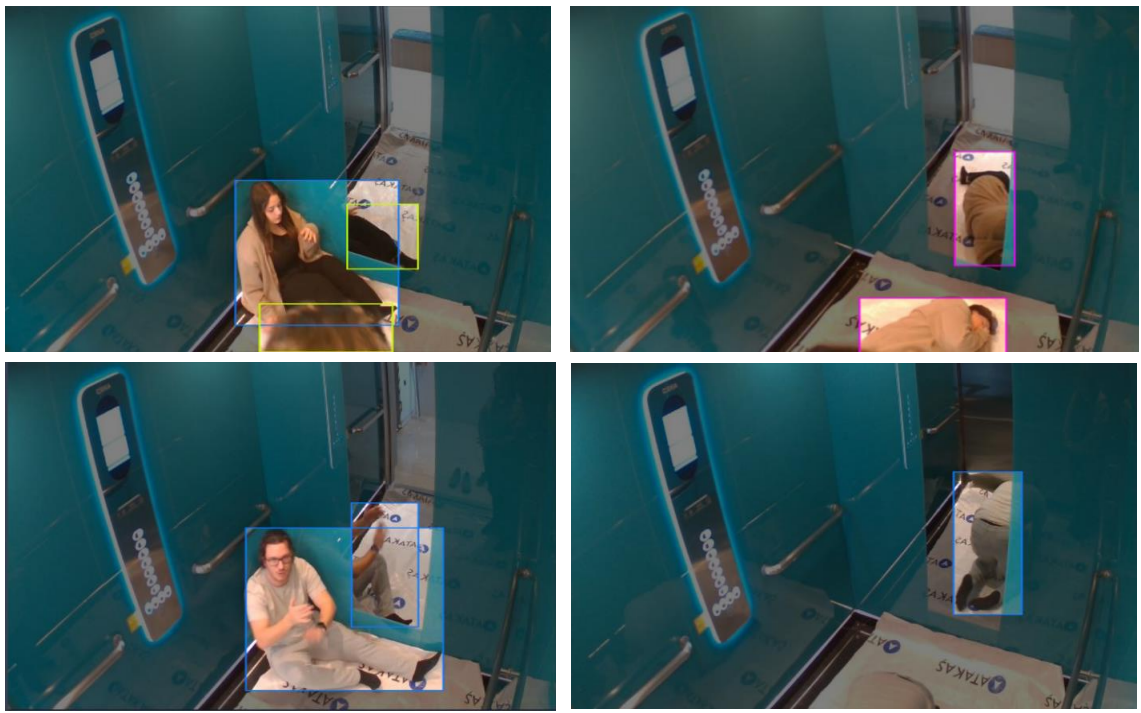

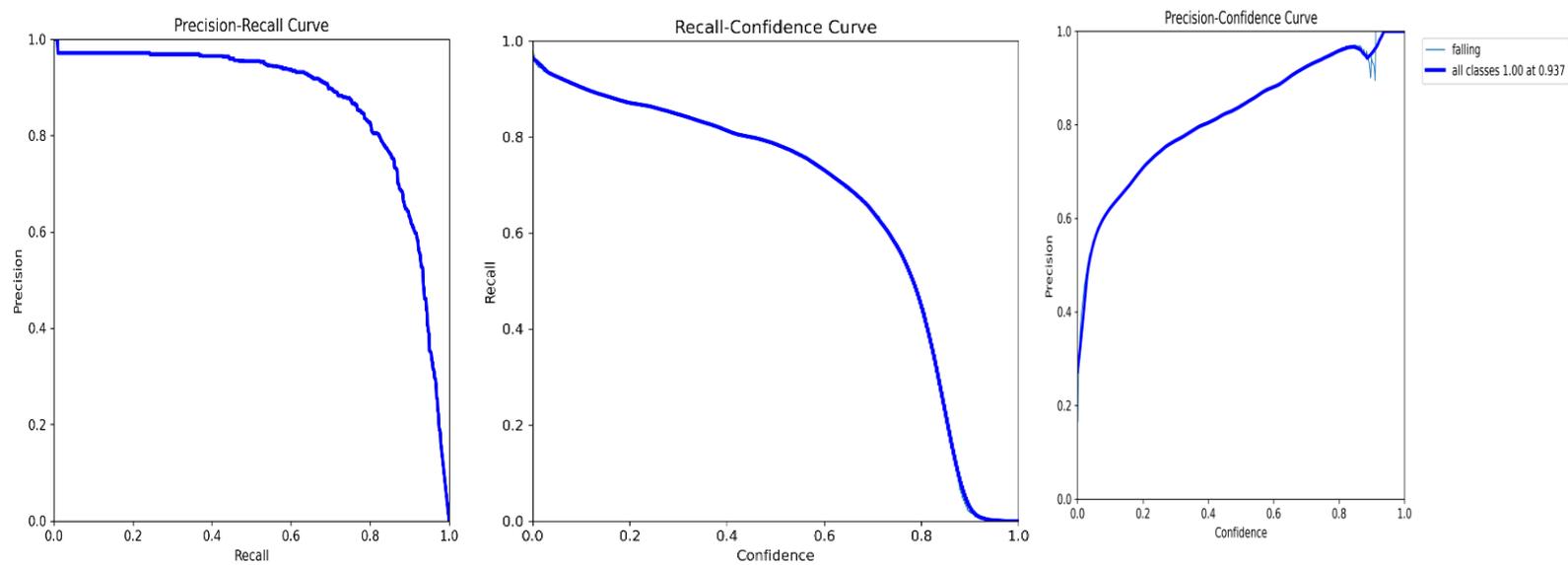

Fig2. Performance for YOLOv8n Model

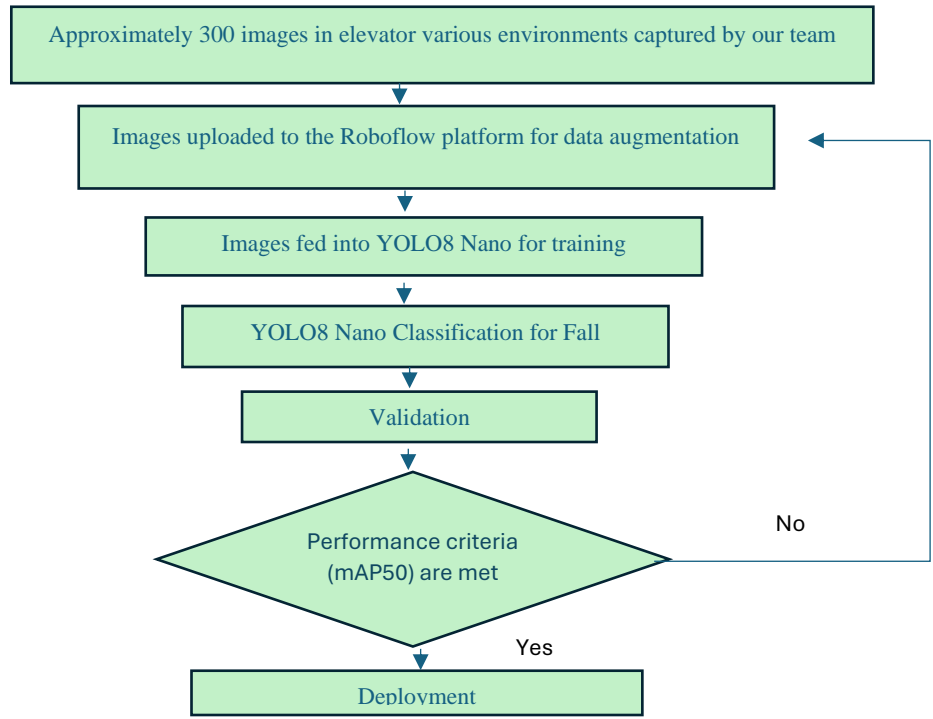

Fig.3. Flowchart of fall detection

**Conclusion**

While elevators are commonly used in high-rise buildings, their enclosed nature poses several safety risks. In crisis situations, such as a fall, intelligent surveillance camera systems can analyze video footage and alert staff to take immediate action. In this study, we developed an artificial intelligence system to detect fall incidents in elevator environments. Despite the challenges such as light reflection, posed by the confined space of an elevator cab when capturing healthy portraits, our findings presents a significant potential to detecting human falls. The YoloV8 Nano model achieved an accuracy of 85% precision and 82% recall, demonstrating that such systems can accelerate emergency response and potentially save lives. These performance results are highly competitive compared to feature-fusion methods, which are more costly in terms of practicality and the duration of the experimental process.

However, the study has some limitations and areas for improvement. Firstly, more real-world data is needed to enhance the model's effectiveness. Secondly, replacing simulated data with actual scenarios could improve accuracy. Lastly, the system requires optimization for real-time performance. Future studies will focus on expanding the dataset and enhancing performance for real-time applications.

**References**

Cao, Z., Simon, T., Wei, S. E., & Sheikh, Y. (2017). Realtime multi-person 2D pose estimation using part affinity fields. In *2017 IEEE Conference on Computer Vision and Pattern Recognition (CVPR)* (pp. 7291-7299). IEEE. https://doi.org/10.1109/CVPR.2017.143

Chen, W., Jiang, Z., Guo, H., & Ni, X. (2020). Fall detection based on key points of human-skeleton using OpenPose. *Symmetry, 12*(5), 744. https://doi.org/10.3390/sym12050744

Chen, Y., Zhao, Q., Fan, Q., Huang, X., Wu, F., & Qi, J. (2023, November). Falling behavior detection system for elevator passengers based on deep learning and edge computing. In *Journal of Physics: Conference Series* (Vol. 2644, No. 1, p. 012012). IOP Publishing. https://doi.org/10.1088/1742-6596/2644/1/012012

Hu, Y., Wang, Q., Wang, C., Qian, Y., Xue, Y., & Wang, H. (2024). MACNet: A More Accurate and Convenient Pest Detection Network. *Electronics*, *13*(6), 1068.

Huang, Z., Liu, Y., Fang, Y., & Horn, B. K. P. (2018). Video-based fall detection for seniors with human pose estimation. In *International Conference on Universal Village (UV)*. IEEE.

Lin, T. Y., Maire, M., Belongie, S., Hays, J., Perona, P., Ramanan, D., ... & Dollár, P. (2014). Microsoft COCO: Common objects in context. In *European Conference on Computer Vision (ECCV)*. Springer.

Liu, P., Lu, T., Lu, Y., Deng, Y., & Lu, Q. (2014). Fall detection based on MEMS three-axis acceleration sensor. *Chinese Journal of Sensors and Actuators, 27*(4), 570-574.

Liu, S., An, Z., Wang, N., Bai, D., & Yu, X. (2021, June). Research on elevator passenger fall detection based on machine vision. In *IOP Conference Series: Earth and Environmental Science* (Vol. 791, No. 1, p. 012108). IOP Publishing. https://doi.org/10.1088/1755-1315/791/1/012108

Lotfi, A., Albawendi, S., Powell, H., Appiah, K., & Langensiepen, C. (2018). Supporting independent living for older adults: Employing a visual-based fall detection through analyzing the motion and shape of the human body. *IEEE Access, 6*, 70272-70282. https://doi.org/10.1109/ACCESS.2018.2876184

Pereira, G. A. (2024). Fall detection for industrial setups using yolov8 variants. *arXiv preprint arXiv:2408.04605*.

Roboflow. (2024). Computer vision annotation tool. Retrieved from https://roboflow.com

Song, F. (2015). Design and implementation of ZigBee and OpenCV-based intelligent monitoring system for the elderly [Master's thesis, Shanghai Jiao Tong University].

Tang, Y., Xie, N., & He, J. (2019). Design and implementation of fall detection algorithm for the elderly based on three-axis acceleration sensor. *Microcomputer Applications, 35*(2), 42-44.

Tong, L., Fan, C., Peng, Z., Wei, C., Sun, S., & Han, J. (2024). WTBD-YOLOv8: An Improved Method for Wind Turbine Generator Defect Detection. *Sustainability*, *16*(11), 4467.

Ultralytics. (2024). Model evaluation insights. Retrieved from https://docs.ultralytics.com/guides/model-evaluation-insights/

Wang, B., Yu, J., Wang, K., Bao, X., & Mao, K. (2020). Fall detection based on dual-channel feature integration. *IEEE Access, 8*, 103443-103453. https://doi.org/10.1109/ACCESS.2020.3000806

Wang, S., Cao, X., Wu, M., Yi, C., Zhang, Z., Fei, H., ... & Yang, P. (2023). Detection of Pine Wilt Disease Using Drone Remote Sensing Imagery and Improved YOLOv8 Algorithm: A Case Study in Weihai, China. *Forests*, *14*(10), 2052.

Xu, C., Xu, Y., Xu, Z., Guo, B., Zhang, C., Huang, J., & Deng, X. (2021, September). Fall detection in elevator cages based on XGBoost and LSTM. In *2021 26th International Conference on Automation and Computing (ICAC)* (pp. 1-6). IEEE. https://doi.org/10.1109/ICAC52482.2021.9586883

Yin, Z. (2020). Research on human abnormal behavior recognition technology based on video [Master's thesis, Changchun University of Science and Technology].

Yu, D., Jiang, X., Jin, G., Lei, F., Yang, Z., Gong, G., & Cui, Z. (2024). ISFDNet: Using low-resolution [Incomplete citation].

Yu, M., Rhuma, A., Naqvi, S. M., Wang, L., & Chambers, J. (2012). A posture recognition-based fall detection system for monitoring an elderly person in a smart home environment. *IEEE Transactions on Information Technology in Biomedicine, 16*(6), 1274-1286. https://doi.org/10.1109/TITB.2012.2214786